\documentclass[conference]{IEEEtran}
\IEEEoverridecommandlockouts

\usepackage{cite}
\usepackage{amsmath,amssymb,amsfonts}
\usepackage{algorithmic}
\usepackage{graphicx}
\usepackage{textcomp}
\usepackage{xcolor}
\usepackage{booktabs}
\usepackage{multirow}
\usepackage{subcaption}
\usepackage[hyphens]{url}

\usepackage{hyperref}
\definecolor{darkblue}{rgb}{0,0,0.5}
\hypersetup{colorlinks=true, citecolor=darkblue, linkcolor=darkblue, urlcolor=darkblue}

\newcommand{\ie}{i.e.,\ }
\newcommand{\eg}{e.g.,\ }

\def\BibTeX{{\rm B\kern-.05em{\sc i\kern-.025em b}\kern-.08em
    T\kern-.1667em\lower.7ex\hbox{E}\kern-.125emX}}

\begin{document}

\title{ControBench: An Interaction-Aware Benchmark for\\
Controversial Discourse Analysis on Social Networks}

\author{%
\IEEEauthorblockN{%
Ta Thanh Thuy\textsuperscript{1},
Jiaqi Zhu\textsuperscript{2},
Xuan Liu\textsuperscript{1},
Lin Shang\textsuperscript{3},
Reihaneh Rabbany\textsuperscript{4,5},\\
Guillaume Rabusseau\textsuperscript{4,6},
Lihui Chen\textsuperscript{1},
Zheng Yilun\textsuperscript{1,*},
Sitao Luan\textsuperscript{4,6,*}%
}
\IEEEauthorblockA{%
\textsuperscript{1}Nanyang Technological University, \;
\textsuperscript{2}NVIDIA, \;
\textsuperscript{3}Nanjing University \\
\textsuperscript{4}Mila -- Quebec AI Institute, \;
\textsuperscript{5}McGill University, \;
\textsuperscript{6}University of Montreal \\
\textsuperscript{*}Corresponding author \quad
\texttt{\{tath0001, yilun001\}@e.ntu.edu.sg}, \;
\texttt{luansito@mila.quebec}%
}%
}
%

\maketitle

\begin{abstract}
Understanding how people argue across ideological divides online is important for studying political polarization, misinformation, and content moderation. Existing datasets capture only part of this problem: some preserve text but ignore interaction structure, some model structure without rich semantics, and others represent conversations without stable user-level ideological identity. We introduce ControBench, a benchmark for controversial discourse analysis that combines heterogeneous social interaction graphs with rich textual semantics. Built from Reddit discussions on three topics, Trump, abortion, and religion, ControBench contains 7,371 users, 1,783 posts, and 26,525 interactions. The graph contains user and post nodes connected by semantically enriched edges; in particular, user-comment-user edges encode both a reply and the parent comment that it responds to, preserving local argumentative context. User labels are derived from self-declared Reddit flairs, providing a scalable proxy for ideological identity without manual annotation. The resulting datasets exhibit low or negative adjusted homophily (Trump: -0.77, Abortion: 0.06, Religion: 0.04), reflecting the cross-cutting structure of real-world debate. We evaluate graph neural networks, pretrained language models, and large language models on ControBench and observe distinct performance patterns across topics and model families, especially when ideological boundaries are ambiguous. These results position ControBench as a challenging and realistic benchmark for controversial discourse analysis. All code, data, and processing scripts are publicly available at \url{https://github.com/thuytt0001/controbench}.
\end{abstract}


\section{Introduction}
\label{sec:introduction}

\begin{figure}[t]
\centering
\includegraphics[width=0.85\linewidth]{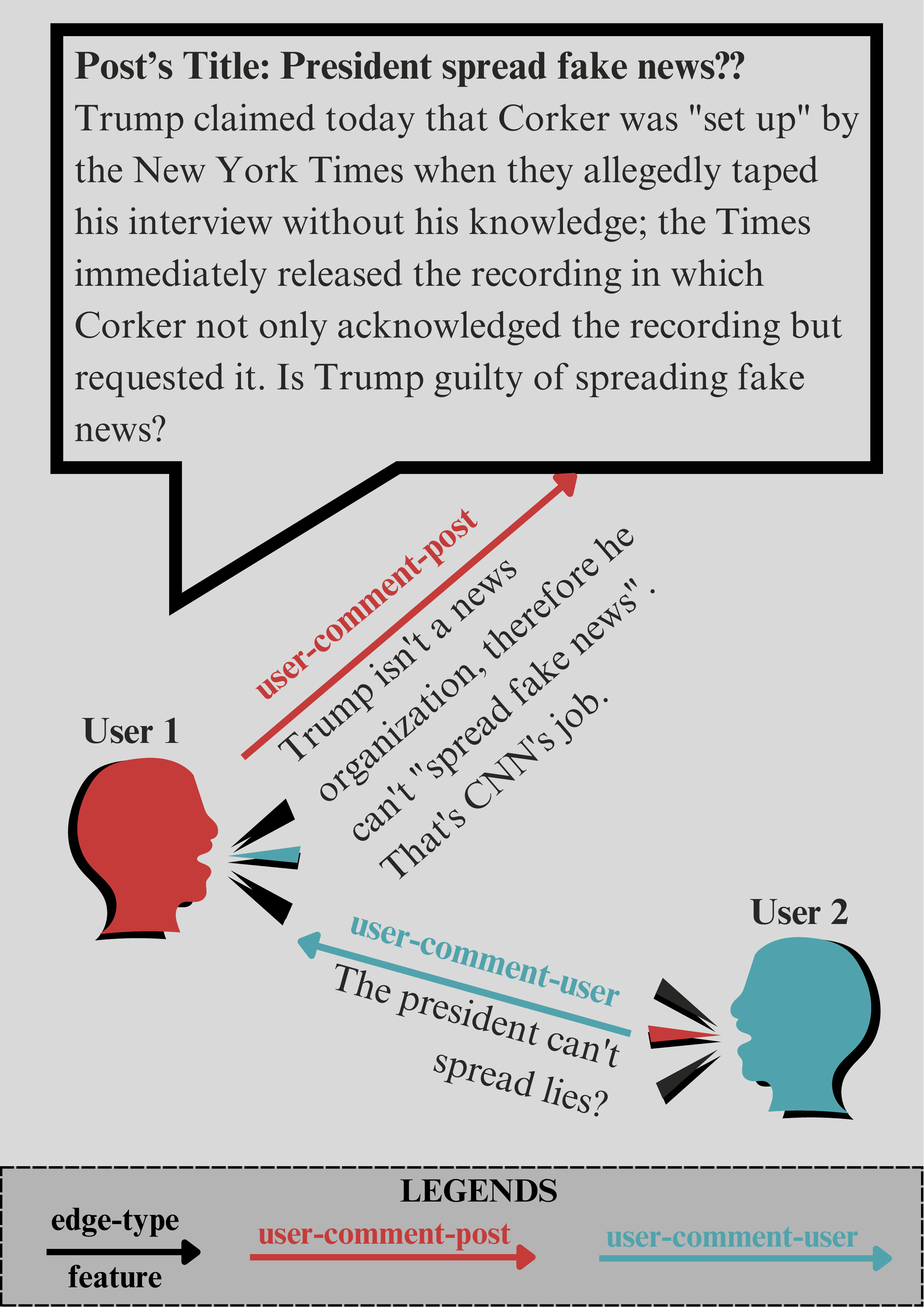}
\caption{Example of interaction in controversial discourse, demonstrating its complexity that requires both semantic and interaction context for correct interpretation.}
\label{fig:satirical_interaction}
\end{figure}

Controversial discourse is a central form of online social interaction. Discussions about politics, abortion, religion, and other value-laden issues shape how people encounter disagreement, form beliefs, and participate in public debate. Understanding such discourse is therefore important for studying political polarization, misinformation, content moderation, and the broader dynamics of digital publics~\cite{vosoughi2018, gillespie2018, bennett2012}. At the same time, controversial discourse is difficult to model: stance is often expressed indirectly through irony, satire, or rhetorical questions~\cite{oxfam2006, dearden1981, bailo2023}, rather than through explicit declarations, as illustrated in Fig.~\ref{fig:satirical_interaction}. As a result, accurate interpretation often requires both semantic understanding and interaction context. 

Existing benchmarks capture only fragments of this problem. Text-attributed graph (TAG) benchmarks~\cite{yan2023comprehensive, li2024tegdb, zhang2024dtgb} provide rich textual information, but they are not designed to model argumentative interactions between people with opposing viewpoints. Conversational stance datasets~\cite{niu2024mtcsd, ding2025zscsd, marreddy2025usdc, zheng2025fanchuan} incorporate reply structure, but typically assign labels at the utterance level, making it difficult to represent stable user-level ideological identity across conversations. Traditional stance detection datasets~\cite{li2021pstance, mohammad2016semeval, glandt2021stance} usually treat posts in isolation, while large-scale social network datasets~\cite{hamilton2017inductive, rozemberczki2021twitch} contain interaction structure but lack the semantic content needed for discourse analysis. Consequently, no existing benchmark jointly models textual semantics, interaction structure, and user-level stance in controversial settings.

To address this gap, we introduce \textbf{ControBench}, a benchmark dataset for controversial discourse analysis that unifies rich textual semantics with interaction-aware heterogeneous graph structure. ControBench is constructed from Reddit discussions on three topics: Trump politics, abortion ethics, and religion. It represents conversations as a heterogeneous graph with two node types (users and posts) and three edge types corresponding to publishing, commenting, and direct user-to-user replies. Importantly, user-comment-user edges carry dual semantic features: one for the reply itself and one for the parent comment being answered. This design preserves local conversational context and allows models to capture not only what users say, but also what they are responding to.

Our main contributions are threefold. \textbf{(1) Dataset Innovation:} We present a benchmark specifically designed for controversial discourse on heterogeneous graphs with rich edge semantics, comprising 7.371 users, 1,783 posts, and 26,525 interactions. User labels are derived from self-declared Reddit flairs, providing stable ideological identity. The datasets exhibit low or negative adjusted homophily, reflecting the cross-cutting and interaction-driven nature of real-world debates. \textbf{(2) Comprehensive Evaluation Framework:} We conduct systematic evaluation across multiple paradigms, including graph-based methods (RGCN~\cite{schlichtkrull2018modeling}, HAN~\cite{wang2019heterogeneous}, HinSAGE~\cite{hamilton2017inductive}, Hetero2Net~\cite{li2025heterophily}, $\mathcal{H}^2$G-Former~\cite{junhong2025heterophily}, H$_2$GCN~\cite{zhu2020beyond}, ACMGNN~\cite{luan2022revisiting}, GCN+GFS~\cite{zhenglet}), pre-trained language models (BERT~\cite{devlin2019bert}, RoBERTa~\cite{liu2019roberta}, SimCSE~\cite{gao2021simcse}, Sentence-BERT~\cite{reimers2019sentence}), and large language models (GPT-4o-mini, Llama-3.1-8B, DeepSeek-V3, DeepSeek-R1, Qwen3-235B, Qwen3-235B-Thinking, and Kimi-K2), alongside strong baselines such as label propagation and COLA. \textbf{(3) Key Empirical Findings:} Our experiments reveal that controversial discourse poses challenges beyond traditional graph or text classification settings. In particular, the combination of heterogeneous structure, cross-ideological interactions, and semantic ambiguity leads to distinct and often inconsistent behaviors across model families, highlighting fundamental limitations of current approaches in modeling real-world social debates.

\section{Related Work}
\label{sec:related_work}

\subsection{Dataset Comparison}

Table~\ref{tab:dataset_comparison} situates ControBench among four lines of work: text-attributed graph benchmarks, conversational stance datasets, isolated-post stance datasets, and general social network datasets.

Text-attributed graph (TAG) benchmarks such as CS-TAG~\cite{yan2023comprehensive}, TEG-DB~\cite{li2024tegdb}, and DTGB~\cite{zhang2024dtgb} enrich nodes and edges with text, but are not designed to model online argumentative interactions. They focus on general graph learning and do not capture how users with opposing beliefs engage in debates.

Conversational stance datasets (MT-CSD~\cite{niu2024mtcsd}, ZS-CSD~\cite{ding2025zscsd}, USDC~\cite{marreddy2025usdc}) incorporate reply structure and multi-turn interactions, but assign stance labels at the utterance level, allowing the same user to take inconsistent labels across conversations. In contrast, ControBench models stable user-level ideological identity via self-declared Reddit flairs, enabling broader analysis beyond stance detection. Additionally, existing datasets use homogeneous reply graphs, whereas ControBench introduces a heterogeneous schema with user and post nodes and three edge types (user-publish-post, user-comment-post, user-comment-user).

A key distinction is the use of dual semantic features on user-comment-user edges, encoding both the reply and its parent comment. This preserves conversational hierarchy and reflects that responses depend on prior context.

Isolated-post stance datasets (SemEval-2016~\cite{mohammad2016semeval}, P-Stance~\cite{li2021pstance}, COVID-19-Stance~\cite{glandt2021stance}) ignore relational structure, while large social network datasets (Reddit~\cite{hamilton2017inductive}, Twitch~\cite{rozemberczki2021twitch}) lack the semantic content required for discourse analysis.

Finally, ControBench exhibits low or negative adjusted homophily (Trump: $-0.7677$, Abortion: $0.0573$, Religion: $0.0441$), indicating predominantly cross-stance interactions. This underscores that ControBench captures the messy, cross-cutting structure of real-world online debate.

\subsection{Methods for Textual Graph Analysis}

Graph neural networks (GNNs) address either heterogeneity or heterophily, but rarely both. Heterogeneous GNNs (RGCN~\cite{schlichtkrull2018modeling}, HAN~\cite{wang2019heterogeneous}, HinSAGE~\cite{hamilton2017inductive}, Hetero2Net~\cite{li2025heterophily}) handle multiple node and edge types but assume homophily. Heterophily-aware GNNs (H$_2$GCN~\cite{zhu2020beyond}, ACMGNN~\cite{luan2022revisiting}, GCN+GFS~\cite{zhenglet}) model cross-class interactions but are restricted to homogeneous graphs. As noted by~\cite{junhong2025heterophily}, jointly heterogeneous and heterophilic graphs remain underexplored, with performance degradation in both families. ControBench lies in this regime, motivating evaluation of methods such as $\mathcal{H}^2$G-Former~\cite{junhong2025heterophily}.

Pre-trained language models (BERT~\cite{devlin2019bert}, RoBERTa~\cite{liu2019roberta}) provide strong semantic representations, while SimCSE~\cite{gao2021simcse} and Sentence-BERT~\cite{reimers2019sentence} offer efficient sentence embeddings without graph structure.

Large language models (LLMs) have shown strong performance via prompting~\cite{wagner2024power, chen2024logic} and multi-agent pipelines such as COLA~\cite{lan2024stance}, but typically ignore graph structure and treat users independently.

Recent work integrates PLMs/LLMs with GNNs. Cascaded approaches (TAPE~\cite{he2024harnessing}) use LLM-generated features for GNNs; iterative approaches (GLEM~\cite{zhaolearning}) jointly optimize both; instruction-based methods (GraphGPT~\cite{tang2023graphgpt}) embed structure into LLM reasoning. However, these methods degrade on heterophilic graphs~\cite{li2025hetgb}. LLM4HeG~\cite{wu2025exploring} addresses this by using LLMs to distinguish homophilic and heterophilic edges, enabling adaptive aggregation~--~particularly relevant for ControBench.

\begin{table*}[t]
\centering
\caption{Comparison of ControBench with representative datasets from existing benchmarks.}
\label{tab:dataset_comparison}
\footnotesize
\setlength{\tabcolsep}{4pt}
\begin{tabular}{lccccc}
\toprule
\textbf{Dataset (Representative)} & \textbf{Nodes/Edges} & \textbf{Text Attributes} & \textbf{Controversial} & \textbf{Social} & \textbf{Heterogeneous} \\
\midrule
\multicolumn{6}{l}{\textit{Text-Attributed Graph Benchmarks}} \\
CS-TAG (ogbn-arxiv-TA)~\cite{yan2023comprehensive} & 169,343 / 1,166,243 & Nodes only & \texttimes & \texttimes & \texttimes \\
TEG-DB (Reddit)~\cite{li2024tegdb} & 478,022 / 676,684 & Nodes \& Edges & \texttimes & \checkmark & \checkmark \\
DTGB (GDELT)~\cite{zhang2024dtgb} & 6,786 / 1,339,245 & Nodes \& Edges & \texttimes & \checkmark & \checkmark \\
\midrule
\multicolumn{6}{l}{\textit{Conversational Stance Datasets}} \\
MT-CSD~\cite{niu2024mtcsd} & 15,876 texts & Nodes only & \checkmark & \checkmark & \texttimes \\
ZS-CSD~\cite{ding2025zscsd} & 17,063 texts & Nodes only & \checkmark & \checkmark & \texttimes \\
USDC~\cite{marreddy2025usdc} & 9,618 texts / 764 threads & Nodes only & \checkmark & \checkmark & \texttimes \\
\midrule
\multicolumn{6}{l}{\textit{Non-graph Social Media Datasets}} \\
SemEval-2016~\cite{mohammad2016semeval} & 4,870 texts & Isolated text & \checkmark & \checkmark & \texttimes \\
P-Stance~\cite{li2021pstance} & 21,574 texts & Isolated text & \checkmark & \checkmark & \texttimes \\
COVID-19-Stance~\cite{glandt2021stance} & 6,133 texts & Isolated text & \checkmark & \checkmark & \texttimes \\
\midrule
\multicolumn{6}{l}{\textit{Social Network Datasets}} \\
Reddit~\cite{hamilton2017inductive} & 232,965 / 114,615,892 & Node features & \texttimes & \checkmark & \texttimes \\
Twitch Gamers~\cite{rozemberczki2021twitch} & 168,114 / 6,797,557 & Node features & \texttimes & \checkmark & \texttimes \\
\midrule
\multicolumn{6}{l}{\textit{ControBench (Ours)}} \\
Trump & 2,067 / 4,975 & Nodes \& Edges & \checkmark & \checkmark & \checkmark \\
Abortion & 2,177 / 11,224 & Nodes \& Edges & \checkmark & \checkmark & \checkmark \\
Religion & 3,127 / 10,326 & Nodes \& Edges & \checkmark & \checkmark & \checkmark \\
\bottomrule
\end{tabular}
\end{table*}

\section{ControBench: A Benchmark for Controversial Discourse Analysis}
\label{sec:controbench}

\subsection{Motivation and Design Principles}

To construct a real-world benchmark dataset for controversial discourse with both textual discussion and social interaction information, ControBench collects content that naturally occurs in Reddit debate communities, where users actively engage across ideological divides. Unlike existing datasets, ControBench provides: (1) rich interaction context showing not just what users say and post, but what they are responding to; (2) multiple relationship types including publishing posts, commenting, and directly replying to arguments; and (3) nuanced stance categories reflecting the actual spectrum of beliefs within controversial topics rather than forcing users into simplified categories.

\subsection{Problem Definition}

We formulate controversial discourse analysis as a node classification problem on heterogeneous graphs with text-attributed edges. Given a heterogeneous graph $\mathcal{G} = (\mathcal{V}, \mathcal{E}, \mathcal{A})$ where:
\begin{itemize}
\item $\mathcal{V} = \mathcal{V}_u \cup \mathcal{V}_p$ represents user nodes and post nodes;
\item $\mathcal{E} = \mathcal{E}_{publish} \cup \mathcal{E}_{comment} \cup \mathcal{E}_{ucu}$ represents three edge types: user-publish-post, user-comment-post, and user-comment-user;
\item $\mathcal{A}$ represents BERT embeddings for textual content.
\end{itemize}

The model learns a function $f: \mathcal{V}_u \rightarrow \mathcal{Y}$ that maps user nodes to stance labels $\mathcal{Y}$. Models must leverage both interacted post content and the conversational context of users to infer their stances, as users have no initial feature representations beyond their interactions and communications. This presents a unique challenge.

\subsection{Data Collection and Construction}

\begin{figure}[t]
\centering
\includegraphics[width=\linewidth]{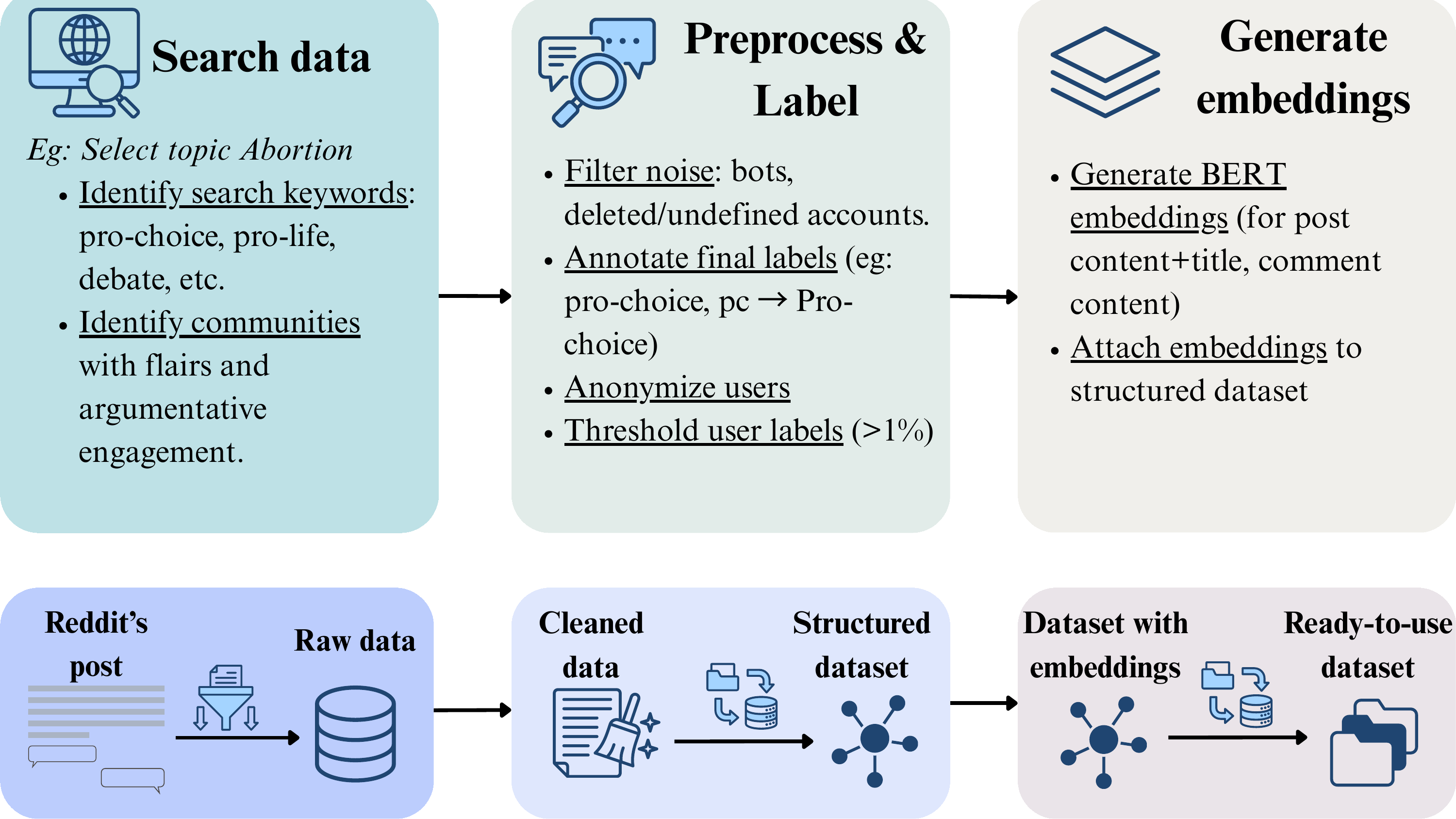}
\caption{ControBench data construction process showing the transformation from Reddit discussions to heterogeneous graphs with rich semantic features.}
\label{fig:data_construction}
\end{figure}

Our dataset collection process is demonstrated in Fig.~\ref{fig:data_construction}.

\textbf{Topic and Community Selection.} We systematically selected three controversial topics and their corresponding Reddit communities using the Reddit Data API~\cite{reddit_api} under academic research terms. Starting with topic-related keywords for Trump, abortion, and religion, such as ``pro-choice'', ``pro-life'', ``atheist'', and ``debate'', we identified relevant subreddits through Reddit search and community recommendations. We prioritized subreddits with identity-related and ideology-related flairs, and based on this grouped the datasets into Trump/abortion and religion settings. The self-identified flairs used in these communities enable systematic stance categorization without requiring manual annotation for every user. We included opposing viewpoints by targeting debate communities (\textit{r/AbortionDebate, r/DebateReligion, r/AskTrumpSupporters}) and supplementing with opposing-spectrum communities when comprehensive debate communities were unavailable.

Our collection spans six subreddits across three topics: Trump (\textit{r/AskTrumpSupporters}), abortion (\textit{r/AbortionDebate, r/prochoice, r/prolife}), and religion (\textit{r/DebateReligion, r/religion}). Data span 2010--2025 with all user identifiers anonymized for privacy compliance. After collecting posts, we extracted user information including all comments on these posts and comments on comments (up to 2-hop depth from each post). We then removed user accounts without flairs, kept users with clear stance indicators, and excluded deleted, bot, or undefined accounts.

\textbf{Content Filtering and Privacy Protection.} We applied systematic filtering to ensure data quality and ethical compliance: (1) removed posts and comments flagged as deleted or removed by moderators; (2) excluded content containing personal information or doxxing attempts; (3) filtered out spam and bot-generated content through automated detection; (4) anonymized all user identifiers using secure hashing methods with no retention of personally identifiable information.

\subsection{Stance Label Assignment}

Our labeling approach addresses controversial discourse complexity through systematic flair-based classification combined with content analysis for edge cases. We leverage the user flair system in Reddit, where users voluntarily self-identify their positions, providing authentic stance indicators. The flair-based label assignment process is as follows:
\begin{enumerate}
\item \textbf{Flair Extraction:} Collected user flairs from all target subreddits.
\item \textbf{Keyword Mapping:} Created and unified comprehensive keyword dictionaries for each topic (\eg ``pro-choice'', ``pc'', ``prochoice'' $\rightarrow$ Pro-Choice).
\item \textbf{Content Verification:} For ambiguous cases, manually analyzed the comment patterns of users for stance confirmation.
\item \textbf{Threshold Application:} Applied a 1\% user representation threshold to remove minority classes, maintaining realistic class distributions while ensuring sufficient samples.
\end{enumerate}
Detailed label mappings are provided in Appendix~\ref{appd:label_assignment}.

\subsection{Stance Label Validation}
\label{sec:iaa}

To assess whether self-declared flairs reflect the ideological identity expressed in users' text, we use Kimi-K2 as an independent text-based annotator: flairs are masked and the model receives the same user content used during evaluation. We compute Cohen's~$\kappa$~\cite{cohen1960, landis1977measurement} between flair-derived labels and model predictions on the 200-user stratified samples (Table~\ref{tab:iaa}).

\begin{table}[t]
\centering
\caption{Stance label validation via Cohen's~$\kappa$ between flair-derived labels and a content-only Kimi-K2 classifier that never observes any stance label during prediction.}
\label{tab:iaa}
\footnotesize
\setlength{\tabcolsep}{5pt}
\begin{tabular}{lcccc}
\toprule
\textbf{Dataset} & \textbf{$\kappa$} & \textbf{Accuracy} &
\textbf{Macro F1} & \textbf{Agreement} \\
\midrule
Trump    & 0.79 & 0.895 & 0.895 & Substantial \\
Abortion & 0.49 & 0.680 & 0.594 & Moderate \\
Religion & 0.46 & 0.530 & 0.547 & Moderate \\
\bottomrule
\end{tabular}
\end{table}

Agreement is substantial for Trump ($\kappa = 0.79$) and moderate for Abortion and Religion, with lower agreement concentrated in inherently ambiguous classes such as Mixed View (F1\,=\,0.22) and Philosophical/Other (F1\,=\,0.30). All three datasets exceed $\kappa > 0.40$, confirming that flair labels carry genuine ideological signal. The moderate values reflect real properties of controversial discourse: users engage strategically across ideological lines, reducing the surface-level distinctiveness of their text relative to their declared stance. This gap between declared identity and expressed text is itself a research target that requires both the flair label and the interaction graph to study.

\subsection{Heterogeneous Graph Structure}

\begin{figure*}[t]
\centering
\includegraphics[width=0.92\linewidth]{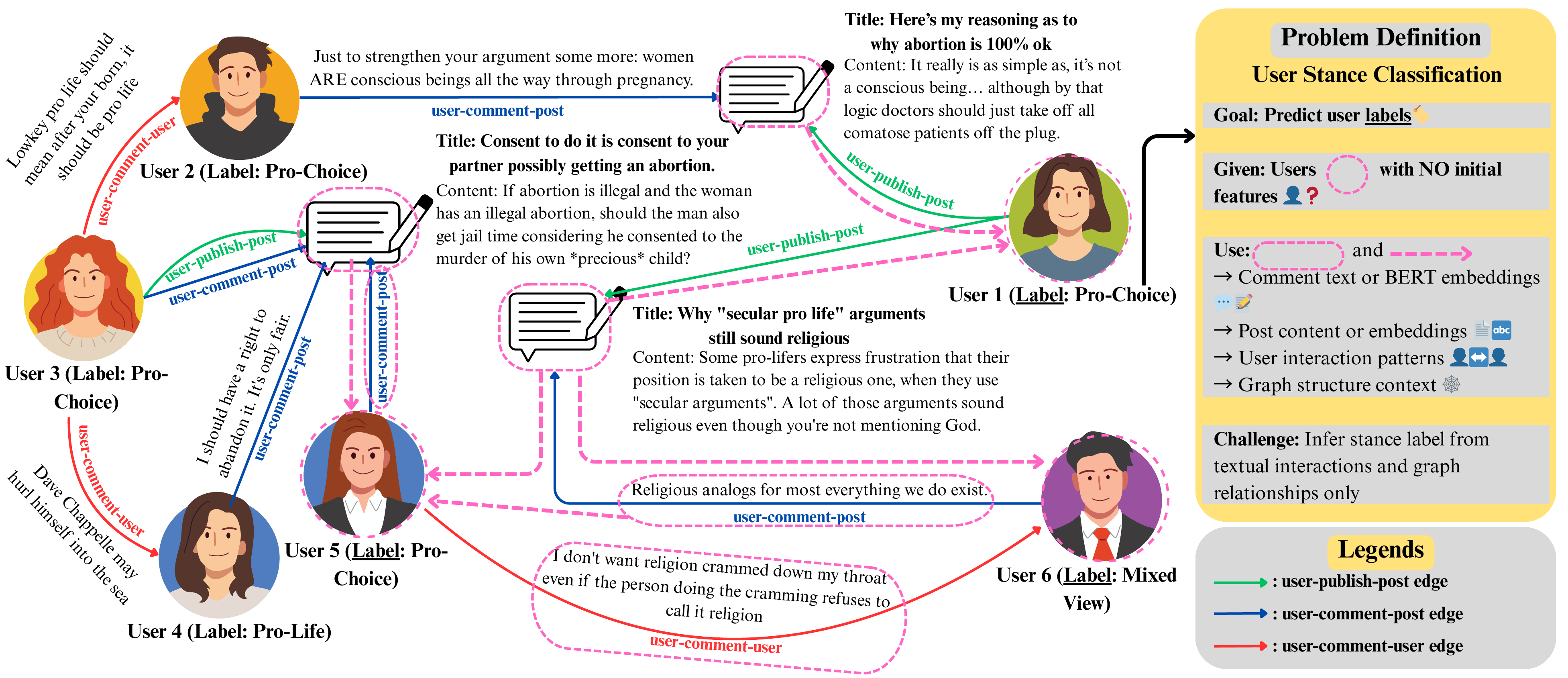}
\caption{ControBench heterogeneous graph structure showing two node types (users, posts) and three semantically-rich edge types that capture publishing, commenting, and direct user interaction patterns. User stances are inferred by leveraging post content and conversation context.}
\label{fig:graph_structure}
\end{figure*}

ControBench captures the full complexity of controversial discourse through a carefully designed graph schema that preserves conversational context and argumentative structure. An example is shown in Fig.~\ref{fig:graph_structure}.

\textbf{Node Types:}
\begin{itemize}
\item \textbf{User nodes:} Represent individuals with stance labels derived from flair analysis. User nodes have no initial features and must be inferred from related post content and conversational context.
\item \textbf{Post nodes:} Contain controversial content with concatenated title and content embeddings (1536-dimensional), available in both raw text and BERT-base-uncased~\cite{devlin2019bert} embedding formats.
\end{itemize}

\textbf{Edge Types with Semantic Features:}
\begin{itemize}
\item \textbf{user-publish-post:} Connect users to posts they created (no features).
\item \textbf{user-comment-post:} Users commenting on posts, enriched with 768-dimensional comment embeddings, available in raw text and BERT~\cite{devlin2019bert} embedding formats.
\item \textbf{user-comment-user:} Direct user interactions through replies, featuring dual semantic features, \ie a 768-dimensional reply content embedding and a 768-dimensional parent comment embedding. The parent comment embedding lets us trace back the original post the conversation thread occurred under. Provided formats are raw text and BERT~\cite{devlin2019bert} embeddings.
\end{itemize}

The dual semantic features on user-comment-user edges represent our key innovation, enabling models to understand both what users say and what they respond to, preserving conversational context crucial for controversial discourse understanding and analysis.

\subsection{Dataset Characteristics and Analysis}

Table~\ref{tab:dataset_stats} presents comprehensive statistics across our three topics, showing the diversity in scale, interaction patterns, and class distributions.

\begin{table}[t]
\centering
\caption{Statistics of ControBench.}
\label{tab:dataset_stats}
\footnotesize
\setlength{\tabcolsep}{6pt}
\begin{tabular}{lccc}
\toprule
\textbf{Statistic} & \textbf{Trump} & \textbf{Abortion} & \textbf{Religion} \\
\midrule
\# Classes & 2 & 3 & 9 \\
\# Posts & 198 & 938  & 647 \\
\# Users & 2,067 & 2,177 & 3,127 \\
\# Edges & 4,975 & 11,224 & 10,326 \\
Avg Degree & 4.39 & 7.21 & 5.47 \\
\bottomrule
\end{tabular}
\end{table}

\textbf{Homophily Analysis.} To quantify interaction patterns in our networks, we compute a variant of the Edge Homophily Ratio (EHR)~\cite{zhu2020beyond} adapted to ControBench's heterogeneous structure. Since only user nodes carry stance labels, we restrict the analysis to user-user neighborhood relationships derived from the graph. For each user $u$, we define their neighbor set $\mathcal{N}(u)$ as the set of distinct users they have interacted with, constructed as follows: (1) if user $u$ comments under a post published by user $v$, then $v \in \mathcal{N}(u)$; (2) if user $u$ directly replies to a comment by user $v$, then $v \in \mathcal{N}(u)$. Multiple interactions between the same pair are counted only once, self-loops are excluded, and post-mediated co-commenters are not treated as neighbors of each other. The EHR is computed over the resulting set of directed user-user pairs:

\begin{equation}
\text{EHR} = \frac{|\{(u,v) : v \in \mathcal{N}(u) \wedge y_u = y_v\}|}{|\{(u,v) : v \in \mathcal{N}(u)\}|}
\end{equation}

where $y_u$ denotes the stance label of user $u$. Values approaching 1 indicate that users predominantly interact with others sharing the same stance, while values near 0 indicate cross-stance interaction. We obtain: 0.1544 (Trump), 0.4759 (Abortion), and 0.2715 (Religion).

Since class sizes differ across datasets, we additionally report the adjusted homophily~\cite{platonov2023characterizing} $\hat{H}$, which accounts for the fact that larger classes naturally produce more same-class edges by chance even under random interaction, making raw EHR scores incomparable across datasets with different class size distributions. All three datasets exhibit near-zero or negative adjusted homophily, reflecting the chaotic, cross-cutting nature of online debate. Trump shows the most strongly heterophilic pattern ($\hat{H} = -0.77$), indicating that users interact predominantly across ideological sides. Abortion ($\hat{H} = 0.06$) and Religion ($\hat{H} = 0.04$) are near-random, where same-stance and cross-stance interactions occur with roughly equal frequency. These patterns stand in stark contrast to the clean homophily assumptions underlying most graph learning benchmarks.

\subsection{Protections and Limitations}

To minimize possible negative impact, we take the following protection actions. \textbf{Privacy Protection:} All user identifiers are anonymized using secure hashing methods with no personally identifiable information retained beyond necessary textual content for research purposes. \textbf{Content Sensitivity:} The dataset contains controversial content requiring careful handling; we provide content warnings and emphasize responsible-use guidelines. \textbf{Responsible Use:} ControBench should be used for advancing scientific understanding of discourse patterns rather than developing systems that could manipulate public opinion or harm democratic processes.

The data we collect may have the following limitations. \textbf{Data Source Limitations:} Reddit user demographics skew toward younger, more educated, primarily English-speaking populations, potentially limiting generalizability. \textbf{Labeling Methodology:} Stance labels assigned through flair analysis lack formal inter-annotator agreement validation, though the flair-based approach provides authentic self-identification; some users may misreport flairs or use ironic/trolling labels. \textbf{Temporal Considerations:} We assume user labels remain static across the collection period (2010--2025), which may not reflect actual opinion evolution.

\section{Experiments}
\label{sec:experiments}

\subsection{Experimental Setup}

\textbf{Evaluation Metrics.} We report Macro-F1 and Micro-F1 scores. We employ a unified train/validation/test split (60\%/20\%/20\%) with stratification across all methods, ensuring consistent class distributions in each split. For PLMs, the split defines training, validation, and held-out evaluation sets in the standard supervised way. For GNNs, we follow a \textbf{transductive} protocol: a single merged graph over all users is constructed, with loss computed on training nodes, early stopping on validation nodes, and test nodes evaluated once without label supervision. We choose the transductive setting to enable message passing across all available nodes and edges.

\textbf{Hyperparameter Search.} For training-based methods (GNNs and PLMs), we conduct hyperparameter search using validation set performance. For GNNs, we search across learning rates, weight decay, dropout rates, hidden sizes, and number of layers. PLM fine-tuning searches across learning rates, batch sizes, training epochs, weight decay, and maximum sequence lengths. Complete search spaces are detailed in Appendix~\ref{appendix:hyperparameter_searching_range_graph_models}.

\textbf{Reproducibility.} For GNNs and PLMs, all results on the full test set are averaged over three independent seeds (42, 123, 456) with the best hyperparameter configuration fixed from the first seed's validation performance; Table~\ref{tab:results_multiseed} reports mean\,$\pm$\,std across seeds. For LLMs, we evaluate on a stratified 200-user sample per dataset (seed\,=\,42) using temperature\,=\,0.0, making outputs fully deterministic. To enable direct cross-category comparison, we additionally evaluate all GNNs and PLMs on the same 200-user sample; Table~\ref{tab:results_200user} reports these results alongside the LLM numbers.

\subsection{Tested Models}

\textbf{Baselines.} We include two baselines representing distinct paradigms. \textbf{Label Propagation}~\cite{zhu2002learning} is a structure-only method that iteratively diffuses stance labels from labeled to unlabeled nodes through the user interaction graph, with the propagation coefficient tuned on the validation set. \textbf{COLA}~\cite{lan2024stance} is a training-free LLM-based method originally designed for individual tweets. We adapt it to ControBench's user-level setting by aggregating each user's posts, comments, and conversation exchanges as a single input. Detailed analysis is in Appendix~\ref{appendix:cola_analysis}.

\textbf{Pre-trained Language Models.} We evaluate four PLM approaches, each operating on a user's concatenated posts, comments, and conversation replies as a single text document. BERT~\cite{devlin2019bert} and RoBERTa~\cite{liu2019roberta} are fine-tuned end-to-end for sequence classification. SimCSE~\cite{gao2021simcse} and Sentence-BERT~\cite{reimers2019sentence} are used as frozen encoders to produce sentence embeddings, with a logistic regression classifier trained on top.

\textbf{Graph Neural Networks and Graph Transformer.} We evaluate eight architectures: HAN~\cite{wang2019heterogeneous}, RGCN~\cite{schlichtkrull2018modeling}, HinSAGE~\cite{hamilton2017inductive}, Hetero2Net~\cite{li2025heterophily}, H$_2$GCN~\cite{zhu2020beyond}, ACMGNN~\cite{luan2022revisiting}, GCN+GFS~\cite{zhenglet}, and the graph transformer $\mathcal{H}^2$G-Former~\cite{junhong2025heterophily}. All models are initialized with BERT-base-uncased~\cite{devlin2019bert} embeddings: user node features are aggregated from their interaction embeddings (comment text, reply text, and post content via mean pooling), post nodes carry 1536-dimensional concatenated title-content embeddings, and comment edges carry 768-dimensional BERT embeddings. For user-comment-user edges, we use embeddings from replies of both users, enabling models to capture full conversational context. Hyperparameter search covers learning rates, weight decay, dropout rates, hidden sizes, and number of layers, with all selections made on the validation set (Appendix~\ref{appendix:hyperparameter_searching_range_graph_models}).

\textbf{Large Language Models.} We evaluate seven LLMs using one-shot prompting with topic-specific examples: GPT-4o-mini, Llama-3.1-8B-Instruct, DeepSeek-V3, DeepSeek-R1, Qwen3-235B, Qwen3-235B-Thinking, and Kimi-K2. Prompts include user context, post content, and explicit stance classification instructions enhanced with parent-reply interactions for conversation-aware analysis. For the LLM experiments, to reduce noise from many interactions, we deduplicate the graph by retaining at most one edge per unique node pair per relation type before constructing user text representations.

\begin{table}[t]
\centering
\caption{Performance comparison on the same 200-user stratified sample (seed\,=\,42), enabling direct comparison across all method categories. Results show Macro F1\,/\,Micro F1 (\%). Bold denotes the best result within each category per dataset. $^\dagger$Label Propagation is evaluated on the full transductive test set; all other methods use the 200-user sample.}
\label{tab:results_200user}
\resizebox{\linewidth}{!}{%
\begin{tabular}{llccc}
\toprule
\textbf{Category} & \textbf{Method} & \textbf{Trump} & \textbf{Abortion} & \textbf{Religion} \\
\midrule
\multirow{2}{*}{Baselines}
 & Label Propagation$^\dagger$ & 54.93 / 58.33 & 58.32 / 63.13 & 51.72 / 44.90 \\
 & COLA                        & 42.47 / 48.48 & 38.24 / 38.00 & 19.35 / 20.60 \\
\midrule
\multirow{8}{*}{GNNs \& GT}
 & HAN              & \textbf{93.49 / 93.5} & \textbf{78.97 / 81.0} & \textbf{52.81 / 51.0} \\
 & RGCN             & 83.19 / 83.5          & 77.18 / 79.0          & 44.40 / 44.0          \\
 & HinSAGE          & 81.10 / 81.5          & 74.34 / 75.5          & 47.42 / 47.5          \\
 & Hetero2Net       & 81.25 / 81.5          & 63.20 / 66.5          & 16.36 / 23.5          \\
 & $\mathcal{H}^2$G-Former & 76.61 / 77.5  & 74.19 / 75.5          & \phantom{0}9.01 / 17.0 \\
 & H$_2$GCN         & 93.50 / 93.5          & 63.69 / 65.0          & 46.87 / 46.5          \\
 & ACMGNN           & 93.49 / 93.5          & 74.17 / 75.0          & 46.93 / 45.0          \\
 & GCN+GFS          & 91.47 / 91.5          & 72.18 / 73.5          & 43.42 / 43.5          \\
\midrule
\multirow{4}{*}{PLMs}
 & BERT          & \textbf{97.00 / 97.0} & 61.63 / 71.0          & 28.98 / 37.0          \\
 & RoBERTa       & 94.49 / 94.5          & \textbf{63.18 / 72.0} & \textbf{42.47 / 47.5} \\
 & SimCSE        & 84.93 / 85.0          & 62.24 / 63.0          & 36.85 / 37.0          \\
 & Sentence-BERT & 72.98 / 73.0          & 62.52 / 64.0          & 40.85 / 41.0          \\
\midrule
\multirow{7}{*}{LLMs}
 & GPT-4o-mini          & 74.26 / 75.0          & 60.70 / 65.5          & 42.14 / 41.5          \\
 & Llama-3.1-8B         & 54.80 / 60.5          & 50.44 / 50.5          & 25.06 / 30.5          \\
 & DeepSeek-V3          & 86.46 / 86.5          & 61.81 / 71.0          & 49.96 / 48.5          \\
 & DeepSeek-R1          & 85.44 / 85.5          & \textbf{62.01 / 72.0} & 49.35 / 47.5          \\
 & Qwen3-235B           & 85.99 / 86.0          & 58.63 / 67.0          & 48.49 / 46.0          \\
 & Qwen3-235B-Thinking  & 85.48 / 85.5          & 61.62 / 73.0          & 49.92 / 49.5          \\
 & Kimi-K2              & \textbf{89.50 / 89.5} & 59.41 / 68.0          & \textbf{54.72 / 53.0} \\
\bottomrule
\end{tabular}%
}
\end{table}

\begin{table*}[t]
\centering
\caption{GNN and PLM performance on the full test set averaged over three random seeds (42, 123, 456). Results show mean\,$\pm$\,std (\%). SimCSE and Sentence-BERT use frozen encoders with logistic regression, so variance across seeds is zero. Bold denotes the best result within each category per dataset.}
\label{tab:results_multiseed}
\resizebox{\linewidth}{!}{%
\begin{tabular}{llcccccc}
\toprule
 & & \multicolumn{2}{c}{\textbf{Trump}} &
     \multicolumn{2}{c}{\textbf{Abortion}} &
     \multicolumn{2}{c}{\textbf{Religion}} \\
\cmidrule(lr){3-4}\cmidrule(lr){5-6}\cmidrule(lr){7-8}
\textbf{Category} & \textbf{Method}
 & \textbf{Macro F1} & \textbf{Micro F1}
 & \textbf{Macro F1} & \textbf{Micro F1}
 & \textbf{Macro F1} & \textbf{Micro F1} \\
\midrule
\multirow{8}{*}{GNNs \& GT}
 & RGCN
   & 87.94$_{\pm 0.37}$ & 88.65$_{\pm 0.38}$
   & 75.17$_{\pm 1.10}$ & 83.41$_{\pm 0.88}$
   & 37.25$_{\pm 1.17}$ & 49.07$_{\pm 1.65}$ \\
 & HAN
   & \textbf{94.40}$_{\pm 0.34}$ & \textbf{94.62}$_{\pm 0.33}$
   & \textbf{75.99}$_{\pm 0.58}$ & \textbf{84.02}$_{\pm 0.88}$
   & \textbf{45.62}$_{\pm 1.02}$ & \textbf{54.81}$_{\pm 2.24}$ \\
 & HinSAGE
   & 87.87$_{\pm 0.49}$ & 88.65$_{\pm 0.43}$
   & 73.87$_{\pm 1.05}$ & 81.74$_{\pm 0.94}$
   & 38.06$_{\pm 0.67}$ & 48.79$_{\pm 1.40}$ \\
 & Hetero2Net
   & 86.09$_{\pm 0.37}$ & 87.00$_{\pm 0.37}$
   & 66.25$_{\pm 4.84}$ & 76.48$_{\pm 2.95}$
   & 15.08$_{\pm 1.02}$ & 40.09$_{\pm 1.35}$ \\
 & $\mathcal{H}^2$G-Former
   & 85.07$_{\pm 1.67}$ & 86.40$_{\pm 1.30}$
   & 70.47$_{\pm 0.99}$ & 78.71$_{\pm 0.81}$
   & 12.19$_{\pm 0.50}$ & 35.28$_{\pm 3.66}$ \\
 & H$_2$GCN
   & 93.17$_{\pm 1.57}$ & 93.50$_{\pm 1.45}$
   & 57.98$_{\pm 2.62}$ & 66.72$_{\pm 2.62}$
   & 37.96$_{\pm 4.27}$ & 47.41$_{\pm 3.04}$ \\
 & ACMGNN
   & 94.69$_{\pm 1.08}$ & 94.92$_{\pm 1.02}$
   & 71.53$_{\pm 0.90}$ & 77.39$_{\pm 0.77}$
   & 43.91$_{\pm 0.89}$ & 51.66$_{\pm 0.73}$ \\
 & GCN+GFS
   & 91.35$_{\pm 0.77}$ & 91.79$_{\pm 0.72}$
   & 68.23$_{\pm 1.53}$ & 74.61$_{\pm 1.04}$
   & 37.24$_{\pm 1.09}$ & 47.13$_{\pm 1.16}$ \\
\midrule
\multirow{4}{*}{PLMs}
 & BERT
   & \textbf{94.63}$_{\pm 0.63}$ & \textbf{94.87}$_{\pm 0.61}$
   & 58.87$_{\pm 4.57}$          & 79.45$_{\pm 0.98}$
   & 41.63$_{\pm 4.74}$          & 64.35$_{\pm 1.36}$ \\
 & RoBERTa
   & 93.24$_{\pm 0.51}$ & 93.56$_{\pm 0.51}$
   & \textbf{63.21}$_{\pm 2.22}$ & \textbf{80.47}$_{\pm 0.88}$
   & \textbf{46.11}$_{\pm 1.26}$ & \textbf{64.03}$_{\pm 0.81}$ \\
 & SimCSE
   & 84.06$_{\pm 0.00}$ & 84.79$_{\pm 0.00}$
   & 52.50$_{\pm 0.00}$ & 59.20$_{\pm 0.00}$
   & 31.39$_{\pm 0.00}$ & 38.05$_{\pm 0.00}$ \\
 & Sentence-BERT
   & 75.08$_{\pm 0.00}$ & 75.85$_{\pm 0.00}$
   & 52.30$_{\pm 0.00}$ & 59.36$_{\pm 0.00}$
   & 35.17$_{\pm 0.00}$ & 41.22$_{\pm 0.00}$ \\
\bottomrule
\end{tabular}%
}
\end{table*}

\subsection{Performance Analysis}

Tables~\ref{tab:results_200user} and~\ref{tab:results_multiseed} present our results. Table~\ref{tab:results_200user} enables direct cross-category comparison by evaluating all methods on the identical 200-user stratified sample; Table~\ref{tab:results_multiseed} reports full-test-set stability across three seeds for GNNs and PLMs.

\textbf{Task complexity is the primary reason for performance differences across all method categories.} All trained methods achieve high performance on Trump (2 classes; best GNN 93.49\%, PLM 97.00\%, LLM 89.50\% on the 200-user sample) but drop sharply on Religion (9 classes), indicating that the number of classes strongly affects performance.

\textbf{Combining interaction structure with text provides clear gains on Abortion.} GNNs jointly leveraging the interaction graph and BERT embeddings achieve the highest macro F1 on Abortion: HAN reaches 75.99\% $\pm$ 0.58 (full test) and 78.97\% (200-user), well above the best PLM and LLM. This may indicate that user interaction networks contain stance information not captured by text alone.

\textbf{Graph structure alone remains highly informative on Religion.} Label Propagation, which uses only the interaction graph without any text, achieves 51.72\% macro F1 on Religion (full test set), exceeding both the best PLM (RoBERTa, 46.11\% $\pm$ 1.26) and the best GNN (HAN, 45.62\% $\pm$ 1.02). On Abortion, LP (58.32\%) remains competitive with PLMs (best 63.21\%), though GNNs outperform it substantially. On Trump, LP (54.93\%) falls well below trained models. These results show that structural signals are particularly valuable in heterophilic, multi-class settings.

\textbf{No single method category dominates across all datasets.} On the 200-user sample, PLMs achieve the best result on Trump (BERT 97.00\%), GNNs on Abortion (HAN 78.97\%), and LLMs on Religion (Kimi-K2 54.72\%). On the full test set, the top GNN (ACMGNN, 94.69\% $\pm$ 1.08) and top PLM (BERT, 94.63\% $\pm$ 0.63) are statistically indistinguishable on Trump given overlapping confidence intervals. On Religion, PLMs (RoBERTa, 46.11\% $\pm$ 1.26) and the best GNN (HAN, 45.62\% $\pm$ 1.02) are again within one standard deviation of each other.

\textbf{GNN performance varies considerably across architectures.} Standard heterogeneous GNNs (HAN, ACMGNN) outperform heterophily-specific architectures ($\mathcal{H}^2$G-Former, Hetero2Net) on every dataset. On Religion, $\mathcal{H}^2$G-Former achieves only 12.19\% $\pm$ 0.50 and Hetero2Net 15.08\% $\pm$ 1.02, far below HAN (45.62\% $\pm$ 1.02). Variance is also notably higher in the multi-class settings: H$_2$GCN shows 37.96\% $\pm$ 4.27 on Religion and Hetero2Net 66.25\% $\pm$ 4.84 on Abortion, indicating sensitivity to initialization. This is consistent with findings~\cite{luan2023graph, luan2024heterophilic, zheng2024missing} that GNNs can struggle more at moderate levels of homophily than at very low levels, and agrees with our analysis in Appendix~\ref{appendix:homophily_vs_performance}, where local homophily does not consistently predict node-level performance. The weaker performance of heterophily-specific architectures may also relate to their design for larger graphs (Appendix~\ref{appd:gnn_implementation}).

\textbf{Among LLMs, larger and more capable models perform better, but reasoning-augmented variants do not consistently improve over their base counterparts.} Kimi-K2 achieves the highest LLM results on Trump (89.50\%) and Religion (54.72\%), while DeepSeek-R1 is strongest on Abortion (62.01\%); the smallest model, Llama-3.1-8B, trails all others by a wide margin (e.g., 54.80\% on Trump, 25.06\% on Religion). Comparing reasoning--base pairs, the effect of explicit reasoning is small and mixed: DeepSeek-R1 (85.44\%) is marginally below DeepSeek-V3 (86.46\%) on Trump but slightly above it on Abortion (62.01\% vs.\ 61.81\%), and Qwen3-235B-Thinking improves over Qwen3-235B on Abortion (61.62\% vs.\ 58.63\%) and Religion (49.92\% vs.\ 48.49\%) yet not on Trump. This indicates that step-by-step reasoning yields no uniform advantage for user-level stance inference. Per-case analysis is in Appendix~\ref{appendix:base_vs_reasoning}.

\textbf{COLA underperforms direct LLM prompting across all datasets.} Despite using GPT-4o-mini as its backbone, COLA achieves only 42.47\%, 38.24\%, and 19.35\% macro F1 on Trump, Abortion, and Religion, falling 31.8, 22.5, and 22.8 points below direct GPT-4o-mini prompting. The multi-agent pipeline does not compensate for the mismatch between COLA's tweet-level design and ControBench's user-level, long-context inputs (Appendix~\ref{appendix:cola_analysis}).

\section{Conclusion and Future Work}
\label{sec:conclusion}

This paper introduces \textbf{ControBench}, a benchmark that integrates heterogeneous graph structures with rich textual semantics for controversial discourse analysis, capturing the complexity of real-world online debates where meaning emerges from both content and conversational context.

All three datasets exhibit low or negative adjusted homophily, confirming that users frequently interact across ideological boundaries rather than within them. Empirically, no single method family dominates across all datasets: GNNs, which jointly leverage interaction structure and text embeddings, achieve the strongest results on Abortion; PLMs perform best on the binary Trump task; and LLMs lead on Religion, where nine-class affiliation classification under near-zero homophily resists both structural and semantic approaches. Notably, Label Propagation, which uses only graph structure without any text, outperforms all PLMs and GNNs on Religion, underscoring that interaction patterns alone carry meaningful ideological signal in cross-cutting debate communities.

These findings reveal that controversial discourse poses challenges beyond traditional graph or text classification settings. The combination of heterogeneous structure, semantic ambiguity, and cross-ideological interaction requires models that can reason over both content and interaction context. Future work should explore interaction-aware graph architectures, hybrid semantic-structural models, spectrum-aware classification for nuanced ideological positions, and temporal dynamics to capture stance evolution. We hope ControBench encourages the development of models capable of handling the complexity of real-world social debate.

\section*{Ethics Statement}

Our data collection and any release of accompanying artifacts are conducted subject to platform policy, applicable law, and explicit authorization for this research use. The data originate from public Reddit discussions accessed through approved research/developer channels. Because the benchmark concerns sensitive topics, we take steps to reduce privacy and misuse risks: user identifiers are irreversibly anonymized, deleted or unavailable accounts are excluded, and we do not attempt to link accounts across platforms or recover real-world identities. We use self-declared community flairs only as coarse, research-only proxies for stance within the studied communities, not as definitive statements about a person's real-world political or religious identity. We do not recommend using ControBench for profiling, targeting, moderation decisions, or any high-stakes downstream application.

\section*{Reproducibility Statement}

We aim to make the construction and evaluation of ControBench transparent and reproducible. The paper and appendix specify the data sources, graph schema, filtering rules, label mapping strategy, feature construction pipeline, model settings, and evaluation protocols used in our study.

\appendix

\section{Label Mappings and Class Distributions}
\label{appd:label_assignment}

Tables~\ref{tab:detailed_class_distributions}, \ref{tab:trump_label}, \ref{tab:abortion_label}, and~\ref{tab:religion_label} summarize the class distributions and flair-to-label mappings used in ControBench. Stance labels are assigned via keyword matching on self-declared flairs, with manual content verification for ambiguous cases.

\begin{table}[h]
\centering
\caption{Detailed class distributions across ControBench datasets.}
\label{tab:detailed_class_distributions}
\footnotesize
\begin{tabular}{lr}
\toprule
\textbf{Class} & \textbf{Users} \\
\midrule
\multicolumn{2}{l}{\textit{Trump Dataset}} \\
Trump Supporter & 1,009 \\
Non-Trump Supporter & 1,058 \\
\midrule
\multicolumn{2}{l}{\textit{Abortion Dataset}} \\
Mixed View & 164 \\
Pro-life & 1,055 \\
Pro-choice & 958 \\
\midrule
\multicolumn{2}{l}{\textit{Religion Dataset}} \\
Non-theistic & 1,413 \\
Christian & 680 \\
Islamic & 264 \\
Jewish & 126 \\
Philosophical/Other & 290 \\
Pagan & 165 \\
Hindu & 59 \\
Satanism & 52 \\
Buddhist & 78 \\
\bottomrule
\end{tabular}
\end{table}

\begin{table}[h]
\centering
\caption{Edge-type distributions across ControBench datasets.}
\label{tab:edge_type_distributions}
\footnotesize
\setlength{\tabcolsep}{6pt}
\begin{tabular}{lrrr}
\toprule
\textbf{Edge Type} & \textbf{Trump} & \textbf{Abortion} & \textbf{Religion} \\
\midrule
comment\_post & 3,442 & 8,937 & 8,923 \\
comment\_user & 1,363 & 1,935 & 1,202 \\
publish      &   170 &   352 &   201 \\
\bottomrule
\end{tabular}
\end{table}

\begin{table}[h]
\centering
\caption{Trump dataset stance labels.}
\label{tab:trump_label}
\scriptsize
\begin{tabular}{ll}
\toprule
\textbf{Label} & \textbf{Flair Keywords} \\
\midrule
Trump Supporter & Nimble Navigator, Trump Supporter \\
Non-Trump Supporter & Non-Trump Supporter, Nonsupporter \\
\bottomrule
\end{tabular}
\end{table}

\begin{table}[h]
\centering
\caption{Abortion dataset stance labels.}
\label{tab:abortion_label}
\scriptsize
\begin{tabular}{ll}
\toprule
\textbf{Label} & \textbf{Flair Keywords} \\
\midrule
Pro-Choice & pro-choice, pc, prochoice, Pro Abortion, Abortion Rights \\
Pro-Life & pro-life, pl, prolife, Anti-Abortion, Abolitionist \\
Mixed View & Morally Against Abortion Legally Pro-Choice, Safe Legal Rare \\
\bottomrule
\end{tabular}
\end{table}

\begin{table}[h]
\centering
\caption{Religion dataset stance labels.}
\label{tab:religion_label}
\scriptsize
\begin{tabular}{ll}
\toprule
\textbf{Label} & \textbf{Flair Keywords} \\
\midrule
Non-theistic & Atheist, Agnostic, Secular, Antitheist, Skeptic, Humanist \\
Christian & Christian, Catholic, Protestant, Orthodox, Baptist, Lutheran \\
Philosophical/Other & Deist, Pantheist, Mystic, Universalist \\
Islamic & Muslim, Islam, Sunni, Shia, Sufi, Ahmadiyya \\
Pagan & Pagan, Wiccan, Druid, Asatru, Heathen \\
Jewish & Jewish, Orthodox Jew, Reform Jew, Conservative Jew \\
Hindu & Hindu, Vaishnavism, Shaivism, Advaita, ISKCON \\
Satanism & Satanist, LaVeyan Satanist, Theistic Satanist \\
Buddhist & Buddhist, Zen, Theravada, Mahayana, Tibetan \\
\bottomrule
\end{tabular}
\end{table}

\section{Implementation and Hyperparameters}
\label{appd:implementation}
\label{appd:gnn_implementation}
\label{appendix:hyperparameter_searching_range_graph_models}

All textual content is encoded with BERT-base-uncased~\cite{devlin2019bert} (max sequence length 256, 768-dim mean-pooled embeddings). Post nodes use 1536-dim concatenated title+content embeddings; edges use single 768-dim embeddings.

\textbf{GNN/PLM hyperparameter search} (validation Macro F1, up to 15 trials per model--dataset pair): learning rates $\{0.001, 0.005\}$; weight decay $\{$1e-4, 5e-4$\}$; dropout $\{0.3, 0.5\}$; hidden sizes $\{128, 256\}$; layers $\{2, 3\}$; HAN/$\mathcal{H}^2$G-Former heads $\{4, 8\}$; GCN+GFS split ratio $\{0.3, 0.5, 0.7\}$. HP search uses 50 epochs (early stopping patience 10); final models retrain for 200 epochs (patience 20) with seed 42, weighted cross-entropy (inverse class frequency), Adam, and ReduceLROnPlateau on validation Macro F1.

\textbf{PLM fine-tuning} (BERT/RoBERTa): learning rates $\{$1e-5, 2e-5, 3e-5, 5e-5$\}$; epochs $\{3, 5, 8\}$; weight decay $\{0.01, 0.1\}$; max length $\{256, 512\}$; FP16 when available. SimCSE/Sentence-BERT use frozen encoders with a logistic-regression head (regularization $C \in \{0.01, 0.1, 1, 10, 100\}$, solver $\in$ \{liblinear, saga, lbfgs\}, class weighting $\in$ \{balanced, None\}).

\textbf{LLMs} are accessed via an OpenAI-compatible endpoint at temperature 0.0, top\_p 1.0. Reasoning models (DeepSeek-R1, Qwen3-235B-Thinking) use a minimum 8,000-token output budget for chain-of-thought. Evaluation uses fixed stratified 200-user samples (seed 42) so that exactly the same users are scored across all models. Prompts present each user's own posts, comments, and parent-reply exchanges (the parent comment text and the user's reply), followed by a label-free interaction count and category-specific hints, instructing a single \texttt{FINAL ANSWER:} line extracted via a multi-stage fallback parser. Crucially, no ground-truth stance label of any neighboring user appears in the prompt; the model classifies each user from content alone.

\emph{Note on heterophily-specific GNNs.} Both $\mathcal{H}^2$G-Former and Hetero2Net were designed for large-scale heterogeneous benchmarks (e.g., ogbn-mag, IMDB) with far more nodes, edges, and relation types than ControBench. Their architectural complexity (multi-head sparse attention with $k$-hop augmentation; multi-path disentangled convolutions with semantic fusion) can over-parameterize on ControBench's smaller, sparser graphs, plausibly explaining their underperformance relative to RGCN and HAN.

\section{Analysis of COLA}
\label{appendix:cola_analysis}

COLA~\cite{lan2024stance} is a multi-agent LLM framework designed for utterance-level stance detection on individual posts. Adapted to ControBench's user-level setting (aggregating each user's content into a single input), it underperforms direct GPT-4o-mini prompting on every dataset, despite sharing the same backbone. We identify four structural causes: (1) \textit{input-length mismatch}~--~COLA's role agents were tuned for sub-280-character tweets, while ControBench inputs can exceed 10{,}000 characters; (2) \textit{agent-disagreement amplification}~--~conflicting expert conclusions push the decision-maker toward hedging labels (Mixed View, Non-theistic); (3) \textit{advocate-stage bias}~--~constructing a case for every candidate class (up to nine on Religion) can rehabilitate wrong hypotheses; and (4) \textit{loss of conversational structure}~--~flattening parent-reply context removes the strongest user-level signal. These results suggest that short-text multi-agent pipelines do not transfer directly to user-level controversial discourse, and that preserving conversational structure matters more than adding reasoning stages.

\section{Homophily vs.\ Performance and Base vs.\ Reasoning Models}
\label{appendix:homophily_vs_performance}
\label{appendix:base_vs_reasoning}

Comparing reasoning and base LLMs, reasoning variants do not uniformly help. On simpler binary patterns (e.g., a user stating a clear pro-choice cutoff), thinking models can over-interpret nuance and misclassify, while base models succeed via direct pattern matching. On more entangled multi-class cases (e.g., a Religion user who cites scholarship without declaring belief), step-by-step reasoning helps separate religious \emph{knowledge} from religious \emph{identity}, favoring the reasoning model.

\section{Satirical Discourse Examples}
\label{appd:satire_examples}

To illustrate why ControBench is challenging for automated stance detection, Tables~\ref{tab:satire_abortion} and~\ref{tab:satire_trump} show two representative cases where literal meaning diverges from intended stance, requiring both semantic and interaction context.

\begin{table}[h]
\centering
\caption{Absurdist satirical argumentation (Abortion dataset).}
\label{tab:satire_abortion}
\scriptsize
\begin{tabular}{p{0.18\linewidth}p{0.74\linewidth}}
\toprule
\textbf{Post} & \textit{``We can stop cancer by making cigarettes legal and accessible for all.''}~--~argues, with deliberately flawed logic, that ``if you are actually prolife\ldots you will support the expansion of tobacco rights.'' \\
\midrule
\textbf{Reply} & ``The difference here being, cigarettes actually harm the person doing said action. Not the best satirical metaphor there.'' \\
\midrule
\textbf{Strategy} & Reductio ad absurdum: an obviously ridiculous cigarette-access parallel mirrors pro-choice arguments to attack perceived inconsistencies in pro-life positions. \\
\midrule
\textbf{Challenge} & The text superficially advocates harmful behavior while actually critiquing the opposing view, so a classifier must separate genuine advocacy from satirical critique. \\
\bottomrule
\end{tabular}
\end{table}

\begin{table}[h]
\centering
\caption{Pointed redirection / Socratic-irony trap (Trump dataset).}
\label{tab:satire_trump}
\scriptsize
\begin{tabular}{p{0.18\linewidth}p{0.74\linewidth}}
\toprule
\textbf{Post} & Asks why 73 former Republican national-security officials endorsed Biden, calling Trump ``dangerously unfit.'' \\
\midrule
\textbf{Quoted} & A supporter dismisses the officials as ``hacks who\ldots fed the industrial war machine\ldots while watching the death of the American manufacturing base.'' \\
\midrule
\textbf{Reply} & ``Can you go explain what parts of this don't fit the current administration?'' \\
\midrule
\textbf{Strategy} & Socratic irony: a seemingly neutral question forces the supporter to confront a contradiction in their own criteria, functioning as a rhetorical trap. \\
\midrule
\textbf{Challenge} & The surface form is a clarifying question, but the intent is adversarial; detecting the embedded ironic positioning requires conversational context. \\
\bottomrule
\end{tabular}
\end{table}

%

\bibliographystyle{IEEEtran}
\bibliography{icdm_2026}

\section*{Acknowledgement}
This project was partially supported by the National Key Laboratory for Novel Software Technology, Nanjing University, through the 2025 Overseas Open Project under Grant KFKT2025A06 and NVIDIA Academic Grant.

\end{document}